\documentclass{article}

\usepackage{arxiv}

\usepackage[utf8]{inputenc} %
\usepackage[T1]{fontenc}    %
\usepackage{hyperref}       %
\usepackage{url}            %
\usepackage{booktabs}       %
\usepackage{amsfonts}       %
\usepackage{nicefrac}       %
\usepackage{microtype}      %
\usepackage{xcolor}         %

\usepackage{graphicx}
\usepackage{easyReview}
\usepackage{caption}
\usepackage{subcaption}
\usepackage{amsmath}
\usepackage{amsfonts}
\usepackage{amsthm}
\usepackage{amssymb}
\usepackage{bbm}
\usepackage{bm}
\usepackage{lipsum}
\usepackage[numbers]{natbib}
\usepackage{stackengine} %
\usepackage{tabularx}
\usepackage{makecell}
\usepackage{booktabs}
\usepackage{diagbox}
\usepackage{enumitem}

\usepackage{wrapfig,lipsum,booktabs}

\usepackage{pgfplots}
\pgfplotsset{compat = 1.9}
\pgfkeys{/pgf/number format/.cd,
         1000 sep={},
         /pgf/number format/fixed,
         /pgf/number format/precision=5 
}
\usepackage{xcolor}
\definecolor{darkgreen}{RGB}{9, 133, 9}
\usepgfplotslibrary{external}
\usepackage{mathtools}

\usepackage{optidef}
\usepackage{mathtools}

\usepackage{xfrac}

\usepackage{algorithm}
\usepackage{algpseudocode}

\usepackage{multirow}%
\usepackage{hhline}%

\usetikzlibrary{decorations.pathreplacing,calligraphy} %

\usepackage{bigints}

\usepackage{relsize,exscale}

\DeclareMathOperator{\Tr}{Tr}
\newcommand{\norm}[1]{\left\lVert#1\right\rVert}

\newtheorem{theorem}{Theorem}[section]
\newtheorem{definition}{Definition}[section]
\newtheorem{remark}{Remark}[section]
\newtheorem{example}[theorem]{Example}
\newtheorem{corollary}[theorem]{Corollary}
\newtheorem{proposition}[theorem]{Proposition}

\usepackage{xspace}
\newcommand{\nystrom}{Nystr\"om\xspace}
\newcommand{\ksvd}{KSVD\xspace}

\usepackage{array}

\usepackage{amsmath}
\usepackage{amssymb}
\usepackage{amsthm}
\usepackage{arydshln}
\usepackage{lipsum}
\usepackage[capitalize,noabbrev]{cleveref}

\theoremstyle{plain}

\title{Nonlinear SVD with Asymmetric Kernels: feature learning and asymmetric Nystr\"om method}

\date{}
\author{
  Qinghua Tao%
  \thanks{Equal contribution.}\hspace{2pt}
  \thanks{Corresponding author.}\hspace{3pt},
  Francesco Tonin%
  \footnotemark[1]\hspace{2pt}
  \footnotemark[2]\hspace{3pt},
  Panagiotis Patrinos and Johan A.K. Suykens\\
Department of Electrical Engineering, ESAT-STADIUS,\\
KU Leuven. Kasteelpark Arenberg 10, B-3001 Leuven, Belgium\\
\texttt{\{qinghua.tao,francesco.tonin,panos.patrinos,johan.suykens\}@esat.kuleuven.be} \\
}

\begin{document}

\maketitle

\begin{abstract}
Asymmetric data naturally exist in real life, such as directed graphs. Different from the common kernel methods requiring Mercer kernels, this paper tackles the asymmetric kernel-based learning problem.
We describe a nonlinear extension of the matrix Singular Value Decomposition through asymmetric kernels, namely \ksvd.
First, we construct two nonlinear feature mappings w.r.t.~rows and columns of the given data matrix. 
The proposed optimization problem  maximizes the variance of each mapping projected onto the subspace  spanned by the other, subject to a mutual orthogonality constraint. Through it Lagrangian,
we show that  it can be solved by the left and right singular vectors in the feature space induced by the asymmetric kernel.
Moreover, we start from the integral equations with a pair of adjoint eigenfunctions corresponding to the singular vectors on an asymmetrical kernel, and extend the  Nystr\"om method to asymmetric cases through the finite sample approximation, which can be applied to speedup the training in \ksvd.
Experiments show that asymmetric \ksvd learns features outperforming Mercer-kernel based methods that resort to symmetrization, and  also verify the effectiveness of  the asymmetric Nystr\"om method.
\end{abstract}

\section{Introduction}\label{sec:intro}
Singular Value Decomposition (SVD) \cite{stewart1993early,strang2006linear,golub2013matrix} performs the factorization of any given matrix $ A \in \mathbb R^{N\times M}$ by two sets of orthonormal eigenbases: $  A =   U \Lambda  V^\top$ with the diagonal matrix $ \Lambda  \geq 0$ of singular values  and the columns of $ U$ and $ V$ being the \emph{left} and \emph{right singular vectors}, respectively.
Principal Component Analysis (PCA) is a method close to but different from SVD. PCA treats the samples  as the rows of $A$ s.t. the set of observations is $\{\bm x_i \in \mathbb R^M\}_{i=1}^N$ with $ A= [\bm x_1, \ldots, \bm x_N]^\top$.
PCA is computed by the eigendecomposition to the symmetric empirical covariance matrix $\frac{1}{N} \sum_{i=1}^N \bm x_i \bm x_i^\top$.
PCA is extended to Kernel PCA (KPCA) by applying a nonlinear feature mapping $\phi\colon \mathbb R^{M} \mapsto \mathbb R^q$ to each sample $\bm x_i$, and then performs PCA in the feature space, i.e. to $\{\phi(\bm x_i) \in \mathbb R^{q}\}_{i=1}^N$ \cite{scholkopf1998}.
KPCA is usually solved by the dual problem involving the eigendecomposition of the symmetric kernel matrix $\hat{K}\bm \alpha=\lambda^2\bm \alpha$, with $\hat{K}_{ij}=\hat{k}(\bm x_i, \bm x_j)$ and $\hat{k}$ being a Mercer kernel \cite{mercer1909}, i.e. $\hat{k}$ is symmetric positive semi-definite, associated with  Reproducing Kernel Hilbert Spaces (RKHS).

Asymmetry exists in many real-world applications, such as directed graphs including citation networks and road networks \cite{khosla2019node,sen2008collective},
sparse approximation \cite{xu2019generalized}, bi-clustering \cite{kluger2003spectral}, the attention in Transformers \cite{vaswani2017,tsai2019,wright2021}, etc.
 While SVD jointly learns both left and right singular vectors in relation to column space and row space for an arbitrary non-square non-symmetric matrix $A$, the  common kernel methods with Mercer kernels~\cite{scholkopf1998,suykens2002least,vapnik1999overview} fail to capture the asymmetry with the employed symmetric kernels.
Although SVD can  process asymmetric  matrices,  SVD lacks flexibility for effective feature learning on more complex data. Hence, similar to the spirit of KPCA mapping the row data into a feature space with the kernel trick for nonlinearity, we investigate:
 
 \emph{how to extend  SVD to a nonlinear form with asymmetric kernel tricks for more flexible learning in feature spaces and meanwhile maintain the asymmetry captured by the two sides of singular vectors?}

Kernel methods additionally suffer from efficiency, as they require to process a kernel matrix quadratic in the sample size.
Many approaches have been proposed to improve the efficiency, among which the Nystr\"om method has been widely applied \cite{NIPS2000_19de10ad, zhang2008improved,zhang2010clustered,NIPS2012_621bf66d,gittens2016revisiting,meanti2020kernel}. The  Nystr\"om method of subsampling arises from the approximation to the integral equation in relation to an eigenvalue problem regarding a symmetric kernel \cite{bakerprenter1981numerical}.
However, the existing Nystr\"om method is derived only in symmetric positive semi-definite (Mercer) kernels. In \cite{nystromsvdnemtsov2016matrix}, it discusses matrix compression  by a Nystr\"om-like method to approximate subparts of the left and right singular vectors, but the symmetric Nystr\"om method is directly applied  to the symmetric submatrix.

In this paper, we derive a nonlinear extension of SVD  through asymmetric Kernels, namely \ksvd, employing two feature mappings w.r.t. to two data sources corresponding to the rows and columns of a given data matrix. 
This in fact resembles the mechanism in SVD, where the left and right singular vectors are jointly learned in relation to  column data and row data (see Theorem \ref{lemma:lanczos}), but in the original input space.
Our \ksvd is related to learning in Reproducing Kernel Banach Spaces (RKBS) \cite{zhang2009reproducing,xu2019generalized,lin2022reproducing,he2022learning} that allow two feature mappings inducing an asymmetric kernel.
Further, we start from the pair of adjoint eigenfunctions in the integral equations related to the SVD described in the  early work of Schmidt \cite{schmidt1907theorie,stewart1993early} and we extend the Nystr\"om method to handle asymmetric kernels, which can be used  to speedup \ksvd training without significant decrease in accuracy of the solution.

The paper is structured as follows. In Section \ref{sec:background}, we first review both asymmetric kernels and classical SVD. In order to extend SVD with asymmetric kernels, we give a formulation of SVD by two data sources scanning rows and columns of the given data matrix through Theorem \ref{lemma:lanczos}. After associating a nonlinear map with each of the two data sources, in Section \ref{sec:ksvd} we formulate an optimization problem in kernel space to jointly maximize the covariance of each feature mapping projected onto the subspace spanned by the other, subject to mutual orthogonality constraints. We derive the \ksvd solution through its Lagrangian, which can be solved by the left and right singular vectors  on an  asymmetric kernel matrix. Later, in Section \ref{sec:nystrom} we derive the  Nystr\"om method for asymmetric kernels through the finite sample approximation to the adjoint eigenfunctions  regarding the left and right singular vectors.
Finally, numerical experiments are presented in Section \ref{sec:exp}, showing that asymmetric \ksvd can learn features outperforming Mercer-kernel based methods that resort to symmetrization and verifying the effectiveness of the asymmetric Nystr\"om method.

\section{Problem Statement}\label{sec:background}

\subsection{Asymmetric Kernels}

Let $\mathcal{X}$ be some input space and $\phi: \mathcal{X} \mapsto \mathcal{H}$ a feature mapping to an RKHS $\mathcal{H}$, with associated kernel $\hat{\kappa}(\cdot, x)=\hat{\phi}(x)$.
In traditional kernel methods, the employed kernel $\hat{\kappa} \colon \mathcal{X} \times \mathcal{X} \mapsto \mathbb{R}$ induced by a \textit{single} feature mapping $\hat{\phi}$ on a \textit{single} data source $\mathcal{X}$ satisfies the Mercer's condition \cite{mercer1909} such that $\hat{\kappa}$ is positive semi-definite and symmetric.
Relaxing this  condition involves studying asymmetric kernels. 

In practice, asymmetric similarity functions are widely used,
 such as  the dot-product attention \cite{vaswani2017} viewed as an asymmetric kernel of an associated RKBS with two input spaces related to queries and keys   in Transformers \cite{wright2021}.
An asymmetric kernel $\kappa \colon \mathcal{X} \times \mathcal{Z} \mapsto \mathbb{R}$ describes a similarity between elements from \emph{two} data sources of $\mathcal{X},\mathcal{Z}$  by \emph{two} feature mappings.
When the data present asymmetric nature, one expects asymmetric kernels to capture more information than symmetric kernels.

\begin{example}[Attention Matrix]
The   attention  in Transformers \cite{vaswani2017} gives the similarity between queries and keys, i.e., $\kappa(\bm t_i, \bm s_j)=\textup{softmax}(\left < W_Q^\top \bm t_i, W_K^\top \bm s_j \right >/\sqrt{d})$, which is asymmetric \cite{tsai2019,wright2021}. 
\end{example}

Despite the utility of asymmetry, the existing kernel methods based on  RKHS, e.g., KPCA,  only deal with symmetric similarities. 
To apply them, one has to resort to symmetrization on an asymmetric similarity matrix $G$,  such as $({G}^\top+{G})/2$, ${G} {G}^\top$, or ${G}^\top {G}$,
which may discard  asymmetric information.
To retain the directionality of data, we consider directly tackling asymmetric kernels $k$.

\subsection{Problem Formulation}
For a general rectangular matrix $A \in \mathbb{R}^{N \times M}$ of rank $r$, its SVD is $A=U \Lambda V^\top$, where $U \in \mathbb{R}^{N \times r}$ and $V \in \mathbb{R}^{M \times r}$  are orthonormal matrices of  the left and right singular vectors, and $\Lambda \in \mathbb{R}^{r \times r}$ is a diagonal matrix with positive entries known as the singular values.
The left singular vectors provide an orthonormal basis of the column space of $A$, while the right singular vectors are an orthonormal basis of the row space of $A$. 
Therefore, first we define two data sources $\mathcal{X}, \mathcal{Y}$, corresponding to rows and columns of the data matrix $A$. 

\begin{definition}[Data sources in \ksvd] \label{def:datasources}
   Given the data matrix $A \in \mathbb{R}^{N \times M}$, two data sources $\mathcal{X}=\{A[i,:] \triangleq \bm x_i\}_{i=1}^N \subset \mathbb{R}^M, \, \mathcal{Z}=\{A[:,j] \triangleq \bm z_j \}_{j=1}^M \subset \mathbb{R}^N$ are constructed by scanning $A$ row-wisely and column-wisely, and are called the row data and the column data, respectively.
\end{definition}

Note that $A$ is the data matrix corresponding to the row data source $\mathcal{X}$ and $A^\top$ is the data matrix corresponding to the column data source  $\mathcal{Z}$.
The following theorem on the SVD of $A$ is of importance.
 
\begin{theorem}[Lanczos \cite{lanczos1958linear}]\label{lemma:lanczos}
Any arbitrary non-zero matrix can be written as $ A=\tilde{ U}\tilde{ \Lambda}\tilde{ V}^\top$, where the matrices $\tilde{ U}, \tilde{ \Lambda}, \tilde{ V}$ are defined by the shifted eigenvalue problem:
\begin{equation}\label{shifted:eigen}
    \begin{aligned}
          A \tilde{ V}& =  \tilde{ U}\tilde{ \Lambda}, \\
         { A}^\top\tilde{ U}& =  \tilde{ V}\tilde{ \Lambda}, 
    \end{aligned}
\end{equation}
where  $\tilde{ U}\in \mathbb R^{N\times r}$ and  $\tilde{ V}\in \mathbb R^{M\times r}$ satisfy $\tilde{ U}^\top\tilde{ U}= I_r$ and $\tilde{ V}^\top\tilde{ V}= I_r$, and $\tilde{ \Lambda}\in \mathbb R^{r\times r}$ is a diagonal matrix with positive numbers.
\end{theorem}

In \cite{suykens2016svd}, it proposes a variational principle to the matrix SVD, connected to the Lanczos decomposition theorem \cite{lanczos1958linear} in the settings of least-squares SVM, relating to the rows and columns of $A$.
We then state the following Corollary, which is of particular interest in our work.   

\begin{corollary} \label{corollary:lanczos}
Let $A \in \mathbb{R}^{N \times M}$ be any non-zero matrix of rank $r$. 
The matrix $A$ can be decomposed as $A=\tilde{U}\tilde{\Lambda}\tilde{V}^\top$, with $\tilde{U}, \tilde{\Lambda}, \tilde{V}$ defined by
\begin{equation*}
\begin{aligned}
    A^\top A \tilde{V} &= A^\top \tilde{U} \tilde{\Lambda} \\
    AA^\top \tilde{U}  &= A \tilde{V} \tilde{\Lambda},
\end{aligned}
\end{equation*}
with positive diagonal matrix $\tilde{\Lambda} \in \mathbb{R}^{r \times r}$ and orthonormal matrices $\tilde{U} \in \mathbb{R}^{N \times r}, \tilde{V}  \in \mathbb{R}^{M \times r}$.
\end{corollary}

The goal of this work is two-fold.
\begin{enumerate}[label=(\roman*)]
    \item Extend SVD with asymmetric kernels through variance maximization across the feature spaces induced by two nonlinear mappings relating to rows and columns (see Section \ref{sec:ksvd}).
    \item Derive the Nystr\"om  method with asymmetric kernels, resulting in computational advantages on larger problems (see Section \ref{sec:nystrom}).
\end{enumerate}

\section{Nonlinear SVD with Asymmetric Kernels} \label{sec:ksvd}

Let the original data matrix be $A \in \mathbb{R}^{N \times M}$ and consider samples $\bm x_i \in \mathcal{X}, \bm z_j \in \mathcal{Z}$  from the two data sources defined in Definition \ref{def:datasources}.
Rather than only working with one nonlinear map of the row data $\bm x_i$ as in KPCA, we apply two feature mappings $\phi\colon \mathbb{R}^M \mapsto \mathbb{R}^p,  \psi\colon\mathbb{R}^N \mapsto \mathbb{R}^p$ to  data sources $\mathcal{X},\mathcal{Z}$:
\begin{align*}
\bm x_i \in \mathbb{R}^M \to \phi(\bm x_i) \in \mathbb{R}^p, \quad 
\bm z_j \in \mathbb{R}^N \to \psi(\bm z_j) \in \mathbb{R}^p,
\end{align*}
where $\mathbb{R}^p$ is the feature space and we assume $\sum_{i=1}^N \phi(\bm x_i) =  \sum_{j=1}^M \psi(\bm z_j) = 0$ with recentering.

\paragraph{Construction of the Subspaces in $\mathbb{R}^p$.} We aim to find a pair of mutually uncorrelated $r$ directions in the feature space $\mathbb{R}^p$ that 
maximize the variance of 
each feature mapping projected onto the $r$-dimensional subspace  spanned by the other. The two sets of directions identify two $r$-dimensional subspaces $\Omega_\phi, \Omega_\psi$ of $\mathbb{R}^p$.
Firstly, we define two projector matrices $A_\phi, A_\psi$ as follows:
\begin{align*} \label{eq:projectors}
A_\phi &\in \mathbb{R}^{p \times r} = [\bm a_{\phi_1}, \dots, \bm a_{\phi_r}], \quad \text{where } \bm a_{\phi_l} \text{ is a vector of the } r\text{-dimensional subspace } \Omega_\phi, \\
A_\psi &\in \mathbb{R}^{p \times r} = [\bm a_{\psi_1}, \dots, \bm a_{\psi_r}], \quad \text{where } \bm a_{\psi_l} \text{ is a vector of the } r\text{-dimensional subspace } \Omega_\psi.
\end{align*}
With $\bm a_{\phi_l}$ belonging to the feature space, we can write it as a linear combination of $\phi(\bm x_i)$ by the representer theorem in RKBS with coefficients $b_{\phi_{li}}$ \cite[Theorem 2]{zhang2009reproducing}:
$
    \bm a_{\phi_l} = \sum_{i=1}^N b_{\phi_{il}} \phi(\bm x_i),
$
 $l=1,\dots, r$, and similarly for $\bm a_{\psi_l}$:
$
    \bm a_{\psi_l} = \sum_{j=1}^M b_{\psi_{jl}} \psi(\bm z_j). 
$
Collecting the coefficients in matrices $B_\phi \in \mathbb{R}^{N \times r}, B_\psi \in \mathbb{R}^{M \times r}$ gives
\begin{equation} \label{eq:coefficcient:matrix}
A_\phi = \Phi^\top B_\phi, \quad A_\psi = \Psi^\top B_\psi,
\end{equation}
where $\Phi \in \mathbb{R}^{N \times p} = [\phi(\bm x_1), \dots, \phi(\bm x_N)]^\top, \Psi \in \mathbb{R}^{M \times p} = [\psi(\bm z_1), \dots, \psi(\bm z_M)]^\top$ are the mapping matrices of the two data sources  in their feature spaces. 
  Note that KPCA \cite{scholkopf1998} only tackles the row data with $\Phi$ and $A_\phi$. More comparisons to symmetric kernel methods are in Supplementary Material.   

\paragraph{Covariances across Feature Spaces.}
We formulate the following two $r \times r$ covariance matrices in the $r$-dimensional subspaces and their corresponding empirical approximations:
\begin{equation}\label{eq:cross:cov}
\begin{array}{rl}
\Sigma_\phi = \text{cov}(\Psi A_\phi, \Psi A_\phi) \approx \widehat{\Sigma_\phi}
= A_\phi^\top \Psi^\top \Psi A_\phi 
= B_\phi^\top \Phi \Psi^\top \Psi \Phi^\top B_\phi 
= B_\phi^\top G G^\top B_\phi,\\
\Sigma_\psi = \text{cov}(\Phi A_\psi, \Phi A_\psi) \approx \widehat{\Sigma_\psi}
= A_\psi^\top \Phi^\top \Phi A_\psi 
= B_\psi^\top \Psi \Phi^\top \Phi \Psi^\top B_\psi 
= B_\psi^\top G^\top G B_\psi.
\end{array}
\end{equation}

The two covariances couple the projections of row data and column data of $A$ in the feature space.
$\Sigma_\phi$ considers projections of the feature mappings of column data $\Psi$ onto the subspace $\Omega_\phi$ spanned by the feature mappings of the row data, and viceversa for $\Sigma_\psi$.

The asymmetric kernel matrix $G \in \mathbb{R}^{N \times M}$ is introduced from the covariances in \eqref{eq:cross:cov} s.t. $G_{ij} = \phi(\bm x_i)^\top \psi(\bm z_j)$.
Instead of explicitly defining the feature mappings $\phi, \psi$, one can simply choose an asymmetric kernel function to achieve an equivalent mapping based on the asymmetric kernel trick.

\begin{definition}[Asymmetric Kernel Trick] \label{def:kernel}
The kernel trick with the kernel function mapping $\kappa: \mathbb R^{M} \times \mathbb R^{N} \mapsto \mathbb R$ can be defined by the inner product of two feature mappings:
\begin{equation}\label{eq:def:kernel}
    \kappa(\bm x, \bm z) = \left< \phi(\bm x), \psi(\bm z) \right>, \quad \forall \bm x \in \mathbb R^{M}, \bm z \in \mathbb R^{N},
\end{equation}
where the output spaces of the feature mappings $\phi$ and $\psi$ are compatible in dimensionality, i.e., $\phi\colon \mathbb R^{M} \mapsto \mathbb R^p$, $\psi\colon \mathbb R^N \mapsto \mathbb R^p$ with $\mathbb R^p$ realizing the compatibility in dimensionality.
\end{definition}

Note that  kernel functions require the two inputs to be compatible in dimensionality, however, $\bm x_i$ and $\bm z_j$ can have different dimensions, as $A$ can be non-square with $N\neq M$. 
In this case, we transform the two data sources into the same dimension (see Remark \ref{rmk:linear:ksvd}).

\paragraph{\ksvd Optimization Problem.}  We formalize the \ksvd problem. Jointly maximizing the sum of the covariances with mutual orthogonality constraints of the projectors gives
\begin{maxi} 
{B_\phi, B_\psi}{\frac{1}{2} \left( \Tr(\widehat{\Sigma_\phi}) + \Tr(\widehat{\Sigma_\psi}) \right)= \frac{1}{2}Tr(B_\phi^\top G G^\top B_\phi) + \frac{1}{2}Tr(B_\psi^\top G^\top G B_\psi) \label{eq:ksvd}}{}{}
\breakObjective{= \frac{1}{2}||G^\top B_\phi||^2_{\rm F} + \frac{1}{2}||G B_\psi||^2_{\rm F}}
\addConstraint{A_\phi^\top A_\psi = B_\phi^\top G B_\psi}{= I_r.}
\end{maxi}

We write the Lagrangian of the \ksvd problem \eqref{eq:ksvd}  in Proposition \ref{prop:dual}, showing that the solutions correspond to Corollary \ref{corollary:lanczos} regarding the asymmetric kernel matrix $G$.

\begin{proposition}[\ksvd Solution] \label{prop:dual}
The solution to the problem
\begin{equation} \label{eq:dual}
\begin{aligned}
G^\top G B_\psi &= G^\top B_\phi \Lambda,\\
G G^\top B_\phi &= G B_\psi \Lambda,
\end{aligned}
\end{equation}
satisfies the same first-order conditions for optimality as the Lagrangian of \eqref{eq:ksvd}, with Lagrange multipliers being the non-zero positive diagonal elements of $\Lambda=\text{diag}\{\lambda_1,\dots,\lambda_r\}$.
\end{proposition}

With Proposition \ref{prop:dual}, we can see that our optimization can be solved by the SVD on the asymmetric kernel matrix $G$, providing the left and singular vectors $B_{\phi}, B_{\psi}$ learned with both nonlinearity and asymmetry introduced on the original data matrix $A$. 
We can  also derive that a special case of our \ksvd with  a specific linear kernel recovers the  original matrix SVD.

\begin{remark}[Linear Kernel in \ksvd to Recover SVD]\label{rmk:linear:ksvd}
Let $\phi(\bm x_i) = C^\top \bm x_i, \, \psi(\bm z_j) = \bm z_j$, with $C \in \mathbb{R}^{M \times N}$. 
If it holds that $ACA=A$,  the matrix SVD is performed \cite{suykens2016svd}, where the  kernel function is defined accordingly as $\kappa(\bm x, \bm z) = \bm x^\top C \bm z$.
In matrix form, we have the linear feature mapping matrices $\Phi =AC$ and $\Psi = A^\top $, and the kernel matrix  $G = \Phi \Psi^\top = ACA = A$; in this case the asymmetric kernel matrix $G$ of \ksvd reconciles to the original data matrix itself, through a linear kernel mapping $\kappa(\bm x, \bm z) = \bm x^\top C \bm z$. Further, we note that 
the condition $ACA=A$ can be satisfied by taking the pseudo-inverse of $A$, i.e. $C=A^\dag$.
\end{remark}

The kernel function $\kappa$ in \ksvd can be chosen asymmetric as the SNE or T kernels \cite{hinton2002stochastic}.
However, the two inputs of kernel functions are commonly required to be compatible in dimensionality, and thus we conduct a compatibility linear transformation on one data source when $N \neq M$, similarly to Remark \ref{rmk:linear:ksvd}.
We note that  using a matrix $C$ as in Remark \ref{rmk:linear:ksvd} can  resolve the dimensionality compatibility  of  $\mathcal{X}$ and $\mathcal{Z}$ by computing $G_{ij} =\tilde{\kappa}(C^\top \bm x, \bm z):= \kappa(\bm x, \bm z) $.
In the special case of a square asymmetric matrix with $N=M$, $C$ is then taken as an identity matrix.

\begin{remark} [Dimensionality Compatibility for Non-square Matrix]\label{rmk:c:matrix}
 The transformation matrix $C$ can be attained by  the pseudoinverse of $A$ \cite{suykens2016svd} (denoted in Experiments as  `a$_0$'), as explained in Remark \ref{rmk:linear:ksvd} for the linear kernel. However, it is computationally expensive and  unstable. We  proposed to consider two more efficient alternatives:

a$_1$) PCA projection on $\bm x_i$: $\min\nolimits_{C}  \lVert A - A  C  C^\top  \rVert^2_{\rm F}$; \quad a$_2$)  randomizing the projection  $C$ for $A$.

The former a$_1$) finds the projection directions  capturing most  information of data samples \cite{pca}
With a$_2$), it is computationally advantageous; driven by the Johnson–Lindenstrauss Lemma \cite{kleinberg1997two,larsen2017optimality}, it shows the main patterns of  data can  be retained with  random linear projections.
\end{remark}

\section{Nystr\"om Approximation for Asymmetric Kernels}
\label{sec:nystrom}

\paragraph{A pair of adjoint eigenfunctions.} In the early work of Schmidt \cite{schmidt1907theorie}, it discusses the treatment of integral equations with an asymmetric kernel for the continuous analogue of SVD  \cite{stewart1993early}. 
With an asymmetric kernel $\kappa(\bm x, \bm z)$,  $\phi(\bm x)$ and $\psi(\bm z)$ satisfying 
\begin{equation}\label{eq:integral:svd}
        \lambda_s  u_s(\bm x) = \int_{\mathcal{D}_x} \kappa(\bm x, \bm z) v_s (\bm z) \, p_z(\bm z) d\bm z,  \quad
    \lambda_s v_s(\bm z) = \int_{\mathcal{D}_z}  \kappa(\bm x, \bm z)u_s (\bm x) \, p_x(\bm x) d\bm x
\end{equation}
are called a pair of adjoint eigenfunctions   corresponding to the eigenvalue $\lambda_s$ with $\lambda_1 \geq\lambda_2 \geq \ldots \geq 0$, where $p_x(\bm x)$ and $p_z( \bm z)$ are the probability densities over $\mathcal{D}_x$ and $\mathcal{D}_z$. 
Note that \cite{schmidt1907theorie} works   with the reciprocal of $\lambda_s$, which is  called  a singular value by differentiating from the eigenvalues of symmetric matrix \cite{stewart1993early}. 
The integral equations \eqref{eq:integral:svd} do not specify the normalization of the adjoint eigenfunctions, which correspond to the left and right singular vectors with finite sample approximation, while in SVD  the singular values are solved as orthonormal. Thus, to correspond the results of the adjoint eigenfunctions to the orthonormal singular vectors in SVD, the scalings determining the norms  are implicitly included in \eqref{eq:integral:svd}.  
Considering the unnormalization of the adjoint eigenfunctions, {we incorporate three scalings  $l_{\lambda_s}, l_{u_s}, l_{v_s}$ for $\lambda_s, u_s(\bm x)$, respectively, $v_s(\bm z)$ into the integral equation \eqref{eq:integral:svd}, such that    $ l_{\lambda_s}\lambda_s  l_{u_s} u_s(\bm x) = \int_{\mathcal{D}_x} \kappa(\bm x, \bm z)  l_{v_s} v_s (\bm z) \, p_z(\bm z) d\bm z$ and $
     l_{\lambda_s}\lambda_s  l_{v_s} v_s(\bm z) = \int_{\mathcal{D}_z}  \kappa(\bm x, \bm z) l_{u_s} u_s (\bm x) \, p_x(\bm x) d\bm x$.}

\paragraph{Nystr\"om approximation for the adjoint eigenfunctions.} Given the i.i.d.~samples  $\{\bm x_1, \ldots, \bm x_n \}$ and $\{\bm z_1, \ldots, \bm z_m \}$, 
from the probability densities   $p_x(\bm x), p_z(\bm z)$ over $\mathcal D_x, \mathcal{D}_z$, 
the  two integral equations in \eqref{eq:integral:svd} over $p_x(\bm x)$ and $p_{z}(\bm z)$ are approximated by   an empirical average:
 \begin{equation}\label{eq:integral:svd:approx}
   \lambda_s u_s(\bm x)  \approx \dfrac{l_{v_s}} {m  l_{\lambda_s}  l_{u_s} }\sum\nolimits_{j=1}^m \kappa(\bm x, \bm z_j)  v_s (\bm z_j), \quad
  \lambda_s  v_s(\bm z)  \approx   \dfrac{l_{u_s}} {n l_{\lambda_s}   l_{v_s}}\sum\nolimits_{i=1}^n \kappa(\bm x_i, \bm z) u_s (\bm x_i),
\end{equation}
where $s=1, \ldots, r$, which corresponds to the rank-$r$ compact SVD on a kernel through the  shifted eigenvalue problem of Theorem \ref{lemma:lanczos}:
\begin{equation}\label{eq:integral:svd:motivate}
       G^{(n, m)} V^{(n, m)}  = U^{(n, m)}\Lambda^{(n, m)}, \quad
        (G^{(n, m)})^\top {U}^{(n, m)} =  {V}^{(n, m)}\Lambda^{(n, m)},
\end{equation}
where $G^{(n, m)} \in \mathbb R^{n\times m}$  is the asymmetric kernel matrix with  entries $G_{ij} = \kappa(\bm x_i, \bm z_j)$ and $r \leq \min \{n, m\}$, $V^{(n, m)} = [\bm v^{(n, m)}_1, \ldots, \bm v^{(n, m)}_r] \in \mathbb R^{m\times r}, U^{(n, m)} = [\bm u^{(n, m)}_1, \ldots, \bm u^{(n, m)}_r] \in \mathbb R^{n \times r}$ are column-wise orthonormal and contain the singular vectors, respectively,  and $\Lambda^{(n, m)} = {\rm{diag}}\{\lambda_1^{(n, m)}, \ldots,\lambda_r^{(n, m)}\}$ denotes the positive singular values.
To match \eqref{eq:integral:svd:approx}  against \eqref{eq:integral:svd:motivate}, we firstly require the scalings on the right-side of the two equations  in  \eqref{eq:integral:svd:approx} to be consistent, i.e.,
${l_{v_s}} / ({m  l_{\lambda_s} l_{u_s}}) \triangleq {l_{u_s}} / ({n l_{\lambda_s}  l_{v_s}})$, which yields  
$
   l_{v_s} = \left (\sqrt{n} / \sqrt{m} \right) l_{u_s}$ and $ {l_{v_s}} / ({m  l_{\lambda_s} l_{u_s}}) \triangleq {l_{u_s}} / ({n l_{\lambda_s}  l_{v_s}}) = \left ({1}/ \sqrt{mn}\right ) l_{\lambda_s}.
$

When running all samplings $\bm x_i$ and $\bm z_j$  in \eqref{eq:integral:svd:approx} to match  \eqref{eq:integral:svd:motivate}, 
we arrive at:
$
    u_s(\bm x_i) \approx \sqrt{\sqrt{mn}l_{\lambda_s}} U^{(n, m)}_{is}$,  $v_s(\bm z_j) \approx \sqrt{\sqrt{mn}l_{\lambda_s}} V^{(n, m)}_{j s}$, $ \lambda_s \approx ({1} /{\sqrt{mn}})l_{\lambda_s}\lambda^{(n, m)}_s.
$
Therefore,  the Nystr\"om approximation to the $s$-th pair of adjoint eigenfunctions with an asymmetric kernel  $\kappa(\bm x, \bm z)$ is obtained:
\begin{equation} \label{eq:nystrom:asymmetric}
\begin{aligned}
    u_s^{(n, m)}(\bm x) & \approx ({\sqrt{\sqrt{mn}l_{\lambda_s}}}/ \lambda_s^{(n, m)} ) \sum\nolimits_{j=1}^m \kappa(\bm x, \bm z_j) V^{(n, m)}_{j s},  \quad s=1, \ldots, r, \\
  v^{(n, m)}_s(\bm z) & \approx ({\sqrt{\sqrt{mn}l_{\lambda_s}}} /\lambda_s^{(n, m)}) \sum\nolimits_{i=1}^n \kappa(\bm x_i, \bm z) U^{(n, m)}_{i s},  \quad s=1, \ldots, r,%
\end{aligned}
\end{equation}
which can also be called the out-of-sample extension for evaluating new samples, where the norm of the singular vectors is up to the scaling $l_{\lambda_s} \in \mathbb R_+$.

\paragraph{Nystr\"om approximation applied to asymmetric kernel matrices.} With the Nystr\"om approximation in  \eqref{eq:nystrom:asymmetric},  we can apply \ksvd to a subset of  data with the sample size $n <N $ and $m< M$  to approximate the adjoint eigenfunctions at all samplings  $\{\bm x_i\}_{i=1}^N$ and  $\{\bm z_j\}_{j=1}^M$. 
We  assume the kernel matrix to approximate from \ksvd is $G \in \mathbb R^{N\times M}$ and  denote $\tilde{\lambda}_s^{(N, M)},  \tilde{\bm u}_s^{(N, M)}$, and $\tilde{\bm v}_s^{(N, M)}$ as the  Nystr\"om approximation of the singular values, and left and right singular vectors of  $G$.
We then utilize  the Nystr\"om method  to approximate the singular vectors of $G$ through  the out-of-sample extension \eqref{eq:nystrom:asymmetric}:
\begin{equation}\label{eq:svd:approx:eigne:sub}
     {\tilde{\bm u}}_s^{(N, M)} = ({\sqrt{{\sqrt{mn}}l_{\lambda_s}}}/ \lambda_s^{(n, m)} )  G_{N,m} \bm v_{s}^{(n, m)}, \quad
     {\tilde{\bm v}}_s^{(N, M)} = ({\sqrt{{\sqrt{mn}}l_{\lambda_s}}}/ \lambda_s^{(n, m)} )G_{n, M}^\top \bm u_{s}^{(n, m)}, 
\end{equation}
with $\tilde{\lambda}_s^{(N, M)} =  ({1}/{\sqrt{mn}l_{\lambda_s}})\lambda^{(n, m)}_s$ for $s=1, \ldots, r$, where  $\bm u^{(n, m)}_{s}, \bm v^{(n, m)}_{s}$ are the left and right singular vectors corresponding to the $s$-th nonzero singular value $\lambda_s^{(n, m)}$ of an $n\times m$ sampled submatrix $ G_{n,m} $, $G_{N,m} \in \mathbb R^{N\times m}$ is  the submatrix  by sampling $m$ columns of $G$, and $G_{n, M}\in \mathbb R^{n\times M}$ is   by sampling $n$ rows of $G$.

\section{Numerical Experiments} \label{sec:exp}

This section consists of two main parts: the evaluations  on \ksvd in asymmetric feature learning and on the asymmetric Nystr\"om method. 
We evaluate the effectiveness of the proposed method in retaining data asymmetry, and it is not to claim that asymmetric kernels are always essential or superior than symmetric ones, which is problem dependent. 
The performance advantage of \ksvd is particularly pronounced   in the cases where  the asymmetric information plays a crucial role.
Experiments are implemented in MATLAB2021b on a PC with a 3.7GHz Intel i7-8700K processor and 64GB RAM.

\subsection{Feature Learning with Directed Data}\label{sec:test:graph}

In directed graphs, the distance  between two nodes has directionality. 
 Directed graphs have wide applications and here we consider three directed graphs, i.e., Cora, Citeseer, and Pubmed \cite{sen2008collective}, which are 
 widely used  as benchmarks. 
\ksvd is compared with its closely related baseline methods, i.e.,  SVD and KPCA, for assessing the efficacy of nonlinearity and asymmetry via \ksvd. More details on the datasets and   setups are in {the Supplementary Material.}

        \begin{table*}[t]
        \caption{Micro-F1 and Macro-F1 scores on the node classification.} \label{tab:node:classify}
        \centering
    \small
        \begin{tabular}{lccccc}
            \toprule
            {Dataset} & {F1 Score ($\uparrow$)} & PCA& KPCA & SVD  & \ksvd  \\	
            \midrule
            \multirow{2}{*}{Cora}       & Micro  & 0.757  & 0.771  & 0.776  &  \textbf{0.792}  \\ 
                                        & Macro  & 0.751& 0.767 & 0.770  & \textbf{0.784}  \\\midrule
            \multirow{2}{*}{Citeseer}   & Micro  & 0.648 & 0.666  & 0.667   &  \textbf{0.678}  \\ 
                                        & Macro  &0.611  & 0.635 & 0.632  & \textbf{0.640}  \\\midrule
            \multirow{2}{*}{Pubmed}    
            & Micro  & 0.765 &0.754  & 0.766&  \textbf{0.773}   \\ 
                                        & Macro & 0.736 &   0.715  &0.738 &  \textbf{0.743}    \\
             \bottomrule
        \end{tabular}
    \end{table*}

\paragraph{Node classification.}\label{sec:node:classification}
We conduct feature extraction with our \ksvd and the compared methods and then perform the classification  based on the extracted features. In directed graphs,  the adjacency matrix is square, hence there is no compatibility issue. With KPCA, we only obtain one set of features, due to the symmetry of its  kernel matrix. With SVD and \ksvd, two sets of features are obtained and all methods keep the first 1000 features, following \cite{khosla2019node,he2022learning}.
We employ an LSSVM classifier with those extracted features as mappings to classify the nodes on Micro-F1 and Macro-F1 scores, as typically used in graphs. 
We apply the RBF kernel for KPCA and  the asymmetric kernel function SNE for \ksvd, where SNE can be seen as an asymmetric extension of RBF.
We conduct 10-fold cross validation for the kernel hyperparameter  searched in the same range, and the average  over 10 runs is reported  in Table  \ref{tab:node:classify},  where ``($\uparrow$)'' indicates larger values for better results.

     In Table  \ref{tab:node:classify}, \ksvd  consistently outperforms PCA, SVD and KPCA,  showing the benefits of deploying nonlinearity upon the asymmetry.
     Compared to KPCA using more flexible nonlinear mappings, the existing linear SVD  already achieves comparable performance, indicating the necessity of exploiting asymmetry in the considered cases. The asymmetric SNE kernel function applied in \ksvd further improves the results of SVD and KPCA, verifying the effectiveness of \ksvd by employing nonlinearity (to SVD) and asymmetry (to KPCA).

    \paragraph{Graph reconstruction.}\label{sec:node:construction}
    Graph reconstruction evaluates how well the extracted features preserve the neighborhood information, which presents the graph structure of node connections. In this task, the adjacency matrix needs to be firstly reconstructed with the given embedding features and then is compared to the ground truth, where the $\ell_1$ and $\ell_2$ norm distances are evaluated.

    \begin{table*}[ht]
        \caption{Different norm distances on the graph reconstruction  task.  } \label{tab:node:reconst}
        \centering
        \small
        \begin{tabular}{lccccc}
            \toprule
    {Dataset} & Measures ($\downarrow$)& PCA& KPCA & {SVD} & {\ksvd}\\
            \midrule
          \multirow{2}{*}{Cora} & $\ell_1$ & 556.0 & 349.0 & 622.0 & \textbf{57.0} \\
        & $\ell_2$& 41.2 & 37.9 &41.7 & \textbf{18.4}  \\
       \midrule
        \multirow{2}{*}{Citeseer} & $\ell_1$ & 138.0 & 46.0 & 176.0 & \textbf{40.0}  \\
        & $\ell_2$& 21.3& 16.0 &24.6 & \textbf{14.3}   \\
       \midrule
         \multirow{2}{*}{Pubmed} & $\ell_1$ & 1937.0 & {171.0} & 1933.0 & \textbf{170.0}\\
        &  $\ell_2$ & 128.0 & 31.9 & 118.1 & \textbf{23.8}\\

            \bottomrule
        \end{tabular}
    \end{table*}
      
    Table \ref{tab:node:reconst}
    evaluates how well the extracted features can preserve the node connection structures in the graph. 
    In this task, \ksvd also achieves the best results. Compared to SVD, our \ksvd greatly improves the performance, illustrating  the significance of introducing nonlinearity in SVD to capture the structural information in the graphs. KPCA achieves distinctively better performances than PCA and SVD in this task, 
   showing that considering the asymmetry alone, e.g., SVD, is not enough for exploring the graph structures and the nonlinearity is of great importance to capture the node structure. The experiments on  directed graphs together demonstrate that both the flexible nonlinearity and  asymmetry exploration are essential  and they can be realized in our  \ksvd.

\subsection{Feature Learning with General Data}\label{sec:general:data}
In this part, rather than the data physically pertaining directed measures, we evaluate \ksvd on general data available on UCI repository \cite{Dua:2019}. The compared methods extract features and then a linear classifier or a regressor is applied for prediction, where the accuracy (ACC) and AUROC metrics are used for classification, while RMSE is  for regression, where  results the test data (20\% of the whole data) are reported. For both KPCA and \ksvd, we take the first 4 left singular vectors of the kernel matrix as the features of each sample. Besides the asymmetric kernel function SNE, we also employ RBF to our \ksvd and note that the resulting kernel matrix $G$ in \eqref{eq:dual} is still asymmetric as the kernel is applied to two different data sources $\mathcal{X}$ and $\mathcal{Z}$, i.e. $\kappa(\bm x_i, \bm z_j)\neq \kappa(\bm x_j, \bm z_i)$. In this experiment, the data matrix is  non-square, so we need the transformation  matrix $C$, which can be attained by alternatives ``$a_0$'', ``$a_1$'' and ``$a_2$'' as in Remark \ref{rmk:c:matrix}.

    \begin{table*}[t!]
        \caption{Downstream task evaluations on extracted features of general UCI datasets.} \label{tab:uci}
        \centering
        \small
        \begin{tabular}{lrccccccc}
            \toprule
             \multirow{2}{*}{Dataset} & \multirow{2}{*}{Metric} & \multirow{2}{*}{KPCA (RBF)}
       & \multicolumn{3}{c}{\ksvd (RBF)} & \multicolumn{3}{c}{\ksvd (SNE)}
       \\ \cmidrule(lr){4-6} \cmidrule(lr){7-9}
     &  &   & $a_0$ & $a_1$ & $a_2$  & $a_0$ & $a_1$ & $a_2$\\	
            \midrule
         \multirow{2}{*}{Diabetes} & ACC ($\uparrow$) &  {0.50} &  {0.50} &  {0.50} & \textbf{0.53} &   {0.50} &  {0.50} & \textbf{0.50}  \\
      &   AUROC ($\uparrow$) & {0.74} &  {0.78} &  {0.78} & {0.76} & 0.77 & \textbf{0.79} &  {0.78}  \\
      \multirow{2}{*}{Ionosphere} &ACC ($\uparrow$) &  {0.75} & {0.73} & \textbf{0.79} & \textbf{0.79} & 0.73 & \textbf{0.79} & \textbf{0.79}  \\
       &  AUROC  ($\uparrow$)&  {0.96} & {0.68} & \textbf{0.98} & 0.93 & {0.67} & \textbf{0.98} & 0.95 \\
           \multirow{2}{*}{Liver} & ACC  ($\uparrow$)& \textbf{0.71} & \textbf{0.71} & \textbf{0.71} & \textbf{0.71} & \textbf{0.71} & \textbf{0.71} & \textbf{0.71}  \\
            &  AUROC  ($\uparrow$) & 0.63 &  {0.70}&  {0.61} & {0.60}& \textbf{0.77} &  {0.60} & {0.60}
            \\
        Cholesterol & RMSE  ($\downarrow$) & \textbf{47.61} &  {49.00} & 49.19 & 49.12 &  {49.00} & 49.26 & 49.58 \\
             Yacht & RMSE  ($\downarrow$) & 14.68 & 14.43 & 14.84 &  \textbf{9.55} &  13.53 & 15.22 &  {9.77} \\
            \bottomrule
        \end{tabular}
    \end{table*}

    In Table \ref{tab:uci}, \ksvd still achieves the best overall performance  over KPCA, showing 
that it is still possible to explore richer information with our \ksvd  in general datasets which are not specified physically with directed measures.    Moreover, we can see that these three alternatives ($a_0$-$a_2$) for the matrix $ C$  all lead to good results. Even with random projections in ``$a_2$'', it still achieve comparably best overall results and outperforms ``$a_0$'' that  computes the pseudo inverse. Thus, the dimensionality compatibility issue in \ksvd can be well resolved with  good prediction results. 
In practice, one can choose these alternatives for $C$ or design in its own way for specific applications.

Nonetheless, we   observe that   the advantages of \ksvd are not always as distinctive as in directed graphs in Section \ref{sec:test:graph}, indicating that less information from to asymmetry is present in the problems. In this regard, we conduct an external evaluation on how much additional information exists due to the asymmetry. We perform density estimation of row data $\mathcal X$ and column data $\mathcal Z$ with $C$ from ``$a_1$'' and then evaluate the KL divergence between these two densities. The KL divergence is attained as: \emph{Diabetes: -9.0, Ionosphere: 457.5, Liver: -509.8, 	Cholesterol: -59.4,	Yacht: 740,} showing that the two data sources in Liver and Cholesterol are less relevant (less ``asymmetric''), which is consistent with our Table \ref{tab:uci}, where KPCA gives comparable results. Therefore, this experiment on general dataset compared to KPCA can also be taken as a knowledge discovery:  how much extra information can be exploited in feature learning by \ksvd tells is related to the asymmetric nature of the data.

\subsection{Evaluation on Asymmetric Nystr\"om Method}
We evaluate the proposed asymmetric Nystr\"om method against other standard solvers on problems of different sizes.
We compare with three common SVD solvers: truncated SVD (TSVD) from the ARPACK library, the symmetric Nystr\"om (Sym.~Nys.) applied to $GG^\top$ and $G^\top G$ employing the Lanczos Method \citep{lehoucq1998} for the SVD subproblems, and randomized SVD (RSVD) \citep{halko2011}.
For all used solvers, we use the same stopping criterion based on achieving a target tolerance $\varepsilon$.
The accuracy of a solution $\tilde{U}=[\tilde{\bm u}_1,\dots,\tilde{\bm u}_r], \tilde{V}=[\tilde{\bm v}_1,\dots,\tilde{\bm v}_r], $ is evaluated as the weighted average 
$\eta = \frac{1}{r} \sum_{i=1}^r w_i ( 1 - |\bm u_i^\top \frac{\tilde{\bm u}_i}{\norm{\tilde{\bm u}_i}}| ) +
\frac{1}{r} \sum_{i=1}^r w_i ( 1 - |\bm v_i^\top \frac{\tilde{\bm v}_i}{\norm{\tilde{\bm v}_i}}| )$,
with $w_i=\lambda_i$ and
 $U=[\bm u_1,\dots,\bm u_r], V=[\bm v_1,\dots,\bm v_r]$  the left and right singular vectors of $G$ from its rank-$r$ compact SVD. 
The stopping criterion for all methods is thus $\eta \leq \varepsilon$. 
This criterion is meaningful in feature learning tasks as the aim is to learn embeddings of the given data, i.e. the singular vectors in \ksvd, rather than approximating the full kernel matrix.
We use random subsampling for all Nystr\"{o}m methods and increase the number of subsamples $m$ to achieve the target $\varepsilon$, employ the SNE kernel, and set $r=20$. In these experiments, we pick $m=n$ as the kernel matrices are square.
More setup details are in Supplementary Material.

\begin{figure}[t!]
\centering
\begin{minipage}{.58\textwidth}
    \centering
    \includegraphics{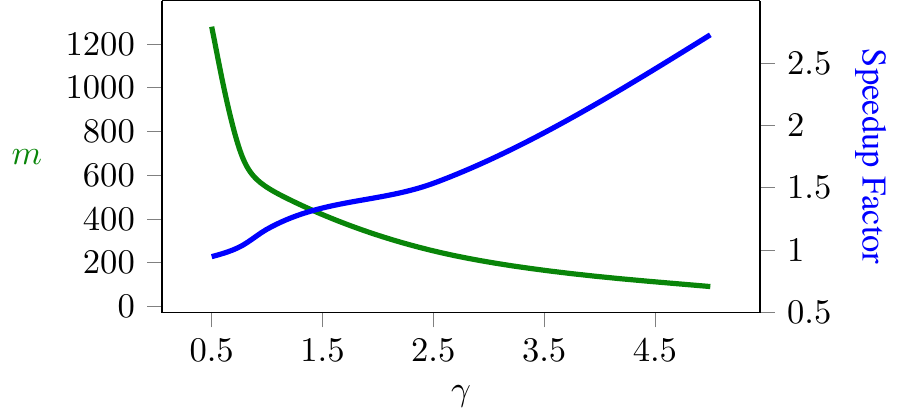}
    \caption{ \textbf{Varying singular spectrum.}
    Number of samples $m$ (green) to achieve a fixed tolerance and the speedup factor w.r.t. RSVD (blue) on Cora when the singular spectrum of $G$ changes (larger $\gamma$ leads to faster decay).
    }
    \label{fig:spectrum}
\end{minipage}%
\hfill
\begin{minipage}{.4\textwidth}
	\centering
    \includegraphics{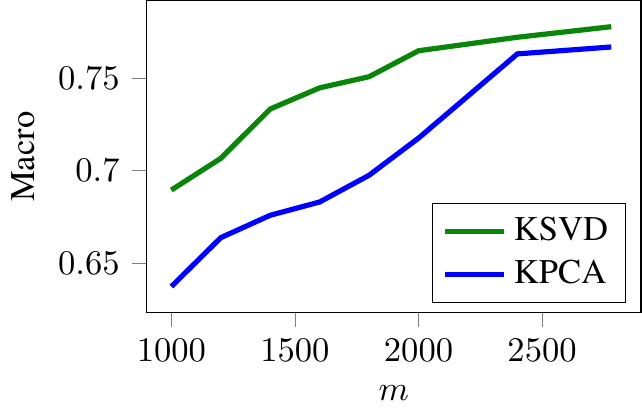}
	\caption{\textbf{Effect of $m$.} Performance at different $m$ for Nystrom applied to \ksvd and KPCA on Cora.}
	\label{fig:nystrom:cora}
\end{minipage}
\end{figure}

Tables \ref{tab:speedup:low} and  \ref{tab:speedup:high} shows the training time on different \ksvd tasks for the tolerance levels $\varepsilon=10^{-1}$ and $10^{-2}$.
We also show the speedup w.r.t. RSVD, i.e. $t^\text{(RSVD)}/t^\text{(Ours)}$, where $t^\text{(RSVD)}, t^\text{(Ours)}$ is the training time of RSVD and our asymmetric Nystrom solver, respectively.
Our solver shows to be the fastest than the compared solvers and our improvement is more significant with larger problem sizes.
Further we consider that a solver's performance may depend on the singular spectrum of the kernel matrix.
We vary the bandwidth $\gamma$ of the SNE kernel on the Cora dataset to assess how the singular value decay  of the kernel matrix  affects the performance, where an increased $\gamma$ leads to spectra with faster decay, and vice versa. 
In Fig.~\ref{fig:spectrum}, we vary $\gamma$ and show the required subsamples $m$ to achieve the given tolerance. We also show the runtime speedup w.r.t.~RSVD.
Our method shows overall speedup compared to RSVD, and our asymmetric Nystr\"om requires significantly fewer subsamples on the matrices with faster decay of the singular spectrum, showing greater speedup w.r.t.~RSVD in this scenario.
In Fig.~\ref{fig:nystrom:cora}, the node classification F1 score (macro) is reported for multiple number of subsamples $m$, where \ksvd employs the asymmetric Nystr\"om method and KPCA uses the symmetric Nystr\"om on the same RBF kernel function. It shows superior performance of the asymmetric method at all considered $m$ without significant decrease in accuracy of the solution due to the subsampling.

    \begin{table*}[t]
        \caption{	
        Runtime for multiple KSVD problems with {higher} tolerance. 
        }
        \label{tab:speedup:low}
        \centering
        \small
        \begin{tabular}{lcccccc}
            \toprule
            \multirow{2}{*}{Task} & \multirow{2}{*}{$N$} & \multicolumn{4}{c}{Time (s) for $\varepsilon=10^{-1}$}  & \multicolumn{1}{c}{Speedup} \\\cmidrule(lr){3-6}
            & & TSVD & RSVD & Sym. Nys & Ours & Factor\\	
            \midrule
            Cora       & 2708  & 0.841  &  0.274  & 0.673   & \textbf{0.160}  & 1.71 \\ 	
            Citeseer   & 3312  & 0.568  &  0.290  & 0.214   & \textbf{0.136}  & 2.14 \\ 	
            PubMed     & 19717 & 9.223  &  4.577  & 44.914  & \textbf{0.141}  & 32.51 \\ 
            \bottomrule
        \end{tabular}
    \end{table*}
    
    \begin{table*}[t]
        \caption{	
        Runtime for multiple KSVD problems with {lower} tolerance. 
        }
        \label{tab:speedup:high}
        \centering
        \small
        \begin{tabular}{lcccccc}
            \toprule
            \multirow{2}{*}{Task} & \multirow{2}{*}{$N$} & \multicolumn{4}{c}{Time (s) for $\varepsilon=10^{-2}$}  & \multicolumn{1}{c}{Speedup} \\\cmidrule(lr){3-6}
            & & TSVD & RSVD & Sym. Nys & Ours & Factor\\	
            \midrule
            Cora       & 2708  & 0.841  &  0.313  & 0.681   & \textbf{0.225}  & 1.39 \\ 	
            Citeseer   & 3312  & 0.568  &  0.396  & 0.425   & \textbf{0.239}  & 1.66 \\ 	
            PubMed     & 19717 & 9.223  &  5.209  & 53.297  & \textbf{0.590}  & 8.83 \\ 
            \bottomrule
        \end{tabular}
    \end{table*}

\section{Conclusion}
This works presents a method for nonlinear SVD employing asymmetric kernels. Two data sources relating to rows of columns of the given data matrix are constructed and transformed through two different feature mappings. In the proposed optimization problem, the variances of the two feature mappings projected onto specifically designed subspaces,  coupling the two data sources, are jointly maximized. 
Through its Lagrangian, we show that this problem corresponds to the SVD of the asymmetric kernel matrix, where the original SVD can be recovered by a  linear kernel function.
 In addition, the asymmetric Nystr\"om method is derived based on the finite sample approximation to the adjoint eigenfunctions, and can be used to speedup the computation of \ksvd.
Numerical results show the potentials of the retained asymmetry and nonlinearity realized in \ksvd and also the effectiveness of the developed asymmetric Nystr\"om method.

\section*{Acknowledgements}
\label{sec:ack}
This work is jointly supported by ERC Advanced Grant E-DUALITY (787960), iBOF project Tensor Tools for Taming the Curse (3E221427), Research Council KU Leuven: Optimization
framework for deep kernel machines C14/18/068, KU Leuven Grant CoE PFV/10/002, and Grant  FWO G0A4917N, EU H2020 ICT-48 Network TAILOR (Foundations of Trustworthy AI - Integrating Reasoning, Learning and Optimization), and the Flemish Government (AI Research Program), and Leuven.AI Institute. This work was also supported by the Research Foundation Flanders (FWO) research projects G086518N, G086318N, and G0A0920N; Fonds de la Recherche Scientifique — FNRS and the Fonds Wetenschappelijk Onderzoek — Vlaanderen under EOS Project No. 30468160 (SeLMA).

\bibliography{bibfile}
\bibliographystyle{unsrt} 

\newpage
\appendix
\section{Further details on the constructed covariances and projections in \ksvd}\label{sec:supp:covariance}

\begin{figure*}[ht]
    \centering  
	\includegraphics[width=0.8\textwidth]{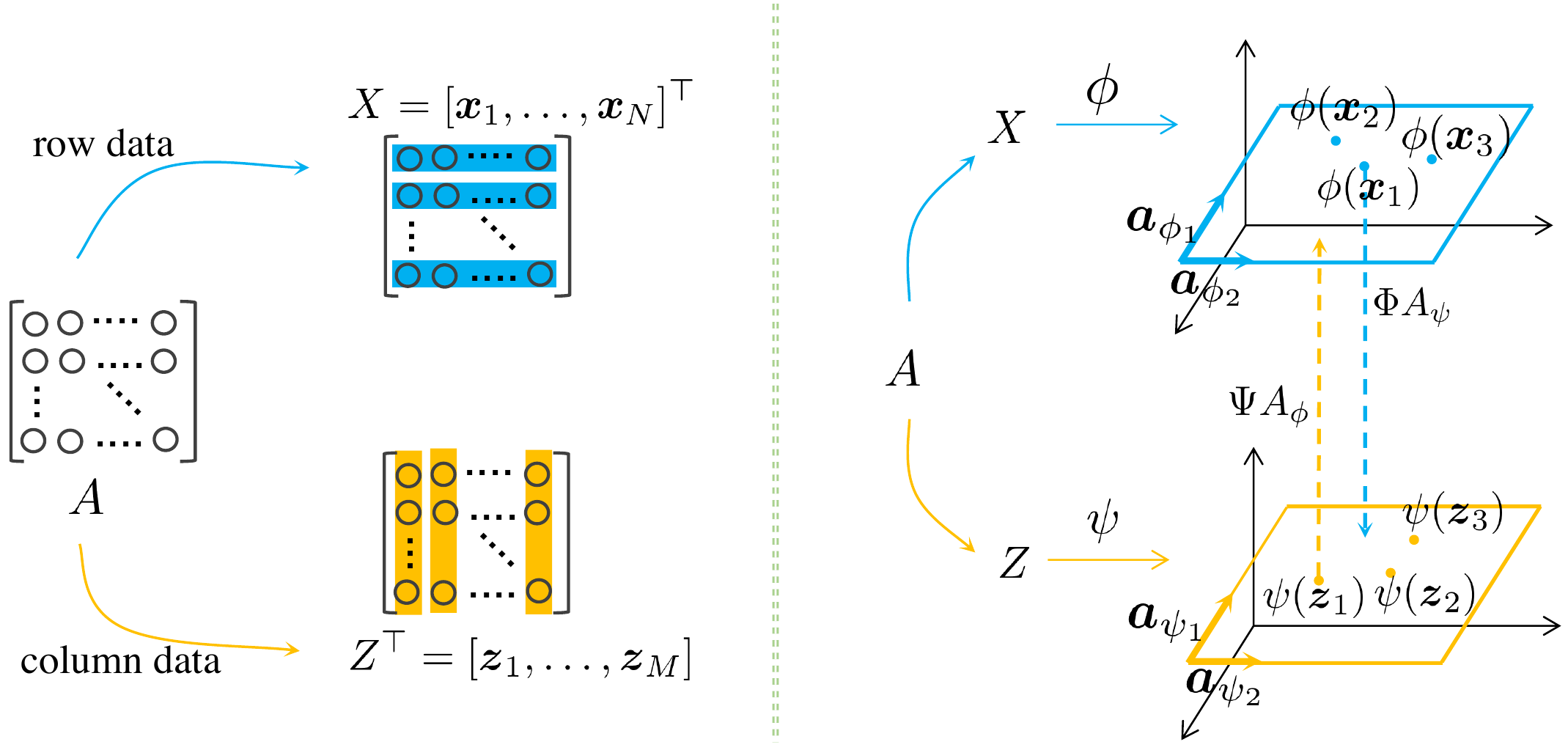}
    \centering  
    \caption{{\bf Schematic illustration of \ksvd.} 
   Given the data matrix $A\in \mathbb R^{N\times M}$, the row data source $\mathcal{X}=\{A[i,:] \triangleq \bm x_i\}_{i=1}^N$ and column data source $\mathcal{Z}=\{A[:,j] \triangleq \bm z_j \}_{j=1}^M$ are constructed, as given in Definition \ref{def:datasources} in the paper. We denote $X \triangleq A$ as the data matrix with samples from  $\mathcal{X}$ and $Z\triangleq A^\top$ as the data matrix with samples from $\mathcal{Z}$. Two feature mappings $\phi$ and $\psi$ are applied to $X$ and $Z$, respectively. Two projectors $A_{\phi}$ and $A_{\psi}$ are constructed through the linear combination of $\phi(\bm x_i)$ and $\psi(\bm z_j)$, respectively, as given in Eq. \eqref{eq:coefficcient:matrix} in the paper. 
   Each mapped data source is then projected onto the subspace spanned by the other and the two covariances given in Eq. \eqref{eq:cross:cov} in the paper are built, where an asymmetric kernel is induced by  the kernel trick $\kappa(\bm x_i, \bm z_j) = \phi(\bm x_i)^\top\psi(\bm z_j)$. 
   The optimization objective of \ksvd aims to find two subpaces (blue and yellow in the Figure), with additional mutual orthogonality constraints, that maximize the constructed covariances in feature space, as given in Eq. \eqref{eq:ksvd} in the paper, so that the maximal information of two data sources can be jointly captured and coupled with each other.
   }
    \label{fig:workflow}
\end{figure*}

To further elaborate on \ksvd, we provide a diagram in Fig. \ref{fig:workflow} exemplifying its mechanism. We also relate its modelling and optimization to KPCA with explanations as follows.

In KPCA, it only considers the row data source corresponding to the rows of $A$, i.e., the data matrix $X=[\bm x_1, \dots, \bm x_N]^\top$ as shown in the left panel in Fig. \ref{fig:workflow}. 
KPCA works with a single feature map $\phi$, while our \ksvd defines two feature mappings $\phi, \psi$ associated with two data sources corresponding to both rows and columns of $A$, i.e., two data matrices $X$ and $Z$.
KPCA aims to find $r$ orthogonal directions in the feature space maximizing the variance of the projections of the feature mappings $\phi(\bm x_i), \, i=1,\dots,N$ onto those directions. 
KPCA works with a single projector matrix $A_\phi$, while \ksvd defines two projector matrices $A_\phi,\, A_\psi$ associated with the two feature mappings, as shown in the right panel in Fig. \ref{fig:workflow}.
KPCA thus performs the eigendecomposition of one covariance matrix $\Sigma^\text{KPCA} = \text{cov}(\Phi A_\phi, \Phi A_\phi)$ \cite{scholkopf1998,nogayama2003}, which, as opposed to \ksvd, does not consider covariances across feature spaces.
Differently, \ksvd considers two covariance matrices 
$\Sigma_\phi = \text{cov}(\Psi A_\phi, \Psi A_\phi)$ and $
\Sigma_\psi = \text{cov}(\Phi A_\psi, \Phi A_\psi)$, 
where each covariance matrix of \ksvd couples the two mapped data sources by considering the projections of one mapped data source onto  the subspace spanned by the other and jointly pursues the maximal variance of the projections.

\section{Further details on the proposed \ksvd and  discussions with related works}\label{sec:supp:related:work}

\subsection{KSVD and discussions}\label{sec:supp:related:work:ksvd}
Our main interest in this work is to  develop new algebraic tools to deal with asymmetry in data, and we focus on the aspect of SVD and its nonlinear extension under the framework of kernel-based learning, namely \ksvd. As explained above in Section \ref{sec:supp:covariance}, \ksvd tackles the given arbitrary data matrix $A$ and attains a kernel matrix in the same size, which is intrinsically different from  KPCA. Through this work, we would also like to convey that although the solutions of PCA and KPCA can be computed numerically by the linear algebra tool of SVD, i.e., the eigendecomposition with a symmetric matrix, PCA is essentially different from SVD, and so is KPCA  from our \ksvd. 

The solution of \ksvd  leads to Corollary 2.3,  with  an asymmetric kernel matrix instead of the given data matrix,  and it closely follows the Lanczos decomposition theorem (Theorem 2.2 \cite{lanczos1958linear}) through the shifted eigenvalue problem interpreting the compact SVD.  Thus, the formulation to the shifted eigenvalue problem is of particular importance in our work.   In \cite{suykens2016svd}, it revisits the matrix SVD with a new variational principle under the setups of least squares support vector machines (LSSVM), where the dual solution leads to a shifted eigenvalue problem regarding the given data matrix.  \cite{suykens2016svd} mainly focuses on the original  (linear) SVD; although it mentions  the possibility with nonlinearity, it does not formalize the   derivations, nor mentions the kernel  tricks or applications. The shifted eigenvalue problem can also trace back to the early work of Schmidt \cite{schmidt1907theorie} that considers the integral equations regarding a pair of adjoint eigenfunctions in the continuous cases with function spaces. Hence, we can see that there can be   multiple frameworks that can lead to a solution in the form resembling a shifted eigenvalue problem either on the given data matrix or an asymmetric kernel matrix as derived in our \ksvd, whereas different goals are  pertained in the addressed scenarios and the methodologies are  also varied with different optimization objectives. 

Moreover, to get the terminology of \ksvd clearer, we   additionally discuss the differences to a few existing works that share some similarities in naming the methodology. In \cite{neto2023kernel},  it considers a new algorithm for SVD that incrementally estimates each set of robust singular values and vectors by replacing the Euclidean norm with the Gaussian norm in the objective.
Different from kernel-based methods, \cite{neto2023kernel} operates in the original space, not in the feature space, where the kernel is only used in the objective for the estimator and the data are not processed with any nonlinearity in the feature space. Despite the similarity in names, the tasks and methodologies in \cite{neto2023kernel} and our \ksvd are intrinsically different. In \cite{he2022learning},  it presents how to apply asymmetric kernels with LSSVMs for supervised classification with both input samples and their labels. In  particular, unlike our \ksvd constructing two data sources, \cite{he2022learning} can only consider one data source (the rows) under the context of its supervised task, exploring the supervised learning for the row data and possibly missing full exploitation of the asymmetry residing in the data. Hence, the data processing, the kernel-based learning scheme, the optimization, and also the task are all different from our \ksvd.

\subsection{Asymmetric Nystr\"om method}\label{sec:supp:related:work:nystrom}

\subsubsection{Details of symmetric Nystr\"om method}\label{sec:supp:sym:nys}
The existing  Nystr\"om method starts from the numerical
treatment of an integral equation with a symmetric kernel  function  $\hat{\kappa}(\cdot, \cdot)$ such that $ \lambda u(x) =  \int_{a}^b \hat{\kappa}( x, z)u ( x) \,d x$, i.e., the continuous analogue to the eigenvalue problem, where the quadrature technique can be applied to formulate the discretized approximation  \cite{bakerprenter1981numerical}. Concerning the more general cases with multivariate inputs, the probability density function and the empirical average technique of finite sampling have been utilized to compute the approximated eigenfunctions that correspond to the eigenvectors \cite{bakerprenter1981numerical,scholkopf1999}. To better illustrate the differences to the established asymmetric Nystr\"om, we provide more details on the  symmetric Nystr\"om  method for reference, based on the  derivations from \cite{NIPS2000_19de10ad}.

Given the i.i.d. samples $\{\bm x_1, \ldots, \bm x_q \}$ from the probability density 
 $p_x(\bm x)$ over $\mathcal D_x$, an empirical average is used to approximate the integral of the eigenfunction with a symmetrick kernel: 
\begin{equation}\label{eq:integral:evd}
    \lambda_s u_s(\bm x) =  \int_{\mathcal{D}_x} \hat{\kappa}(\bm x, \bm z)u_s (\bm x)p_x(\bm x) \,d\bm x \approx \frac{1} {q}\sum\nolimits_{i=1}^q \hat{\kappa}(\bm x, \bm x_i)u_s (\bm x_i),
\end{equation}
where $u_s$ is said to be an eigenfunction of $\hat{\kappa}(\cdot, \cdot)$ corresponding to the eigenvalues with $\lambda_1\geq \lambda_2 \geq \ldots \geq 0$. By running $\bm x$ in \eqref{eq:integral:evd} at $\{\bm x_1, \ldots, \bm x_q \}$, an eigenvalue problem is motivated, such that  $K^{(q)}U^{(q)}=U^{(q)}\Lambda^{(q)}$, where $K^{(q)}\in \mathbb R^{q\times q}$ is the Gram matrix with $K_{ij}^{(q)}=\hat{\kappa}(\bm x_i, \bm x_j)$ for $i,j =1, \ldots, N$, $U^{(q)}=[\bm u^{(q)}_1, \ldots, \bm u^{(q)}_q] \in \mathbb R^{q\times q}$ is column orthonormal and the diagonal matrix $\Lambda^{(q)} \in \mathbb R^{q\times q}$ contains the eigenvalues  such that $\lambda_1^{(q)} \geq \ldots  \geq \lambda_q^{(q)} \geq 0$. In this case, the approximation of eigenvalues and eigenfunction from the integral equation \eqref{eq:integral:evd} arrives at: 
\begin{equation}\label{eq:approx:evd}
 \lambda_s  \approx \frac{\lambda_s^{(q)}}{q}, \quad
     u_s(\bm x_i) \approx \sqrt{q} U^{(q)}_{i,s},
\end{equation}
which can be plugged back to \eqref{eq:integral:evd}, leading to the Nystr\"om approximation  to the $i$-th eigenfunction:
\begin{equation}\label{eq:sym:nystrom:u}
    u_s(\bm x) \approx \dfrac{\sqrt{q}}{\lambda^{(q)}_s}\sum_{i=1}^q \hat{\hat{\kappa}}(\bm x, \bm x_i) U^{(q)}_{i,s},
\end{equation}
with $\forall s: \lambda^{(q)}_s >0$.
With the Nystr\"om technique in \eqref{eq:sym:nystrom:u}, one can use different sampling sets to approximate the integral \eqref{eq:integral:evd}. Thus, given a larger-scale Gram matrix  $K^{(N)}\in \mathbb R^{N\times N}$, for the first $p$ eigenvalues and eigenfunctions, a subset of  training data $q\triangleq n< N$ can be utilized to attain their approximation   at all $N$ points for the kernel matrix $K^{(N)}$ with \eqref{eq:approx:evd}: 
\begin{equation}\label{eq:approx:evd:kernel}
      \tilde{\lambda}^{(N)}_s  \triangleq \dfrac{N}  {n}\lambda_s^{(n)},\quad
     {\tilde{\bm u}}_s^{(N)} \triangleq \sqrt{\dfrac {n}{N}} \dfrac{1}{\lambda_s^{(n)}} K_{N,n} \bm u^{(n)}_{s}, 
\end{equation}
where $ \tilde{\lambda}^{(N)}_s $ and $  {\tilde{\bm u}}_s^{(N)}$ are the Nystr\"om approximation of the eigenvalues and eigenvectos of $K^{(N)}$. Here $ \bm u^{(n)}_{s}$  are  eigenvectors corresponding to the $s$-th   eigenvalues $\lambda_s^{(n)}$ of an $n\times n$ submatrix $K_{n,n}$ and $K_{N,n}$ is the submatrix by sampling $n$ columns of  $K^{(N)}$.%

\subsubsection{Discussions}\label{sec:supp:sym:nys:discuss}
We provide the following take-away messages that help understand the Nystr\"om methods regarding the  eigenvalue problem for symmetric matrices and the  SVD problem for asymmetric matrices.
\begin{enumerate}
    \item \textbf{Integral equations.} The symmetric Nystr\"om method starts from  a single integral equation with a symmetric kernel $\hat{\kappa}(\cdot, \cdot)$, corresponding to an eigenvalue problem in the discretized scenarios \cite{bakerprenter1981numerical,NIPS2000_19de10ad}. Differently, the proposed asymmetric Nystr\"om method deals with an asymmetric kernel ${\kappa}(\cdot, \cdot)$ and starts from a pair of adjoint eigenfunctions, which jointly determine an SVD problem in the discretized scenarios \cite{schmidt1907theorie,stewart1993early}.  In \cite{nystromsvdnemtsov2016matrix}, it discusses the matrix compression task and proposes  a Nystr\"om-like method to general matrices; however, the method is formulated to approximate subparts of the left and right singular vectors, and still applies the symmetric Nystr\"om method  to heuristically  approximate the asymmetric submatrix twice for the corresponding subparts. Hence, the analytical framework of the asymmetric  Nystr\"om method has not been formally formulated yet. In our paper, the explicit rationale of leveraging the Nystr\"om technique is provided for the asymmetric matrices, so that from analytical and practical aspects it becomes viable  to directly apply the asymmetric Nystr\"om method to the cases that pertain the asymmetric nature.
    \item \textbf{Special case with symmetry.} In the derivations on the finite sample approximation, three  scalings $l_{\lambda_s}$, $ l_{u_s}$, and $l_{v_s}$ are introduced to the singular  values $\lambda_s$, right singular vectors $u_s(\bm x)$, and left singular vectors $v_s(\bm z)$ in Eq.  (7) in the paper, for the considerations on their norms; meanwhile the constant coefficients in the two equations in Eq.  (8) in the paper are required to be the same in scalings, where the left and right singular vectors are coupled. 
    In the symmetric  Nystr\"om method, the scaling issue of the approximated eigenfunction does not appear with $ l_{\lambda_s} \lambda l_{u_s} u(\bm x) =  \int \hat{\kappa} (\bm x, \bm z) l_{u_s} u(\bm x)p_x(\bm x) \,d\bm x$, as the scaling $l_{u_s}$  is cancelled out in the two sides of this equation, i.e., \eqref{eq:integral:evd}. Thus, in \eqref{eq:integral:evd} it implicitly sets the scaling  of the eigenvalue as $l_{\lambda_s}=1$  \cite{NIPS2000_19de10ad}, while  in \eqref{eq:approx:evd:kernel} $l_{\lambda_s}$ is set as $1/ N$ in the application of the Nystr\"om method to speedup the eigenvalue problem on a larger  Gram matrix $K^{(N)}$.

    Note that, for feature learning, we only need to find the singular vectors in Eq. (10) in the paper, which are taken as embeddings of the given data for downstream tasks.
    The computation of the singular values can be omitted, so that we can simply implement the scaling through normalization in practice.
    The numerical computation of the approximated kernel matrix is also not necessary for the considered feature learning tasks.

    When considering the special case where the kernel matrix $G$ in \ksvd is square ($ N=M$) and symmetric  ($G = G^\top$), the numbers of samplings to the rows and column are the same ($n=m$),  and the scaling $l_{\lambda_s}$ is set the same, the asymmetric Nystr\"om method boils down to  the existing Nystr\"om method.

\end{enumerate}

\section{Further numerical evaluations}

\subsection{Ablation study}
To further evaluate the effectiveness of the simultaneous nonlinearity and asymmetry introduced in \ksvd, we design the following experiment.
We first make some non-linear encoding in a preprocessing step to the given matrix and then compute SVD, and compare the downstream classification/regression results with our proposed method.
Specifically, we consider polynomial features with degree 2 and then apply SVD.
Correspondingly, our \ksvd employs the polynomial kernel of degree 2.

\begin{table}[H]
    \centering
    \caption{Ablation study on SVD applied after nonlinear preprocessing v.s. \ksvd.
    Higher values ($\uparrow$) are better for AUROC and lower values ($\downarrow$) are better for RMSE.} 
    \label{tab:mytable}
    \begin{tabular}{lccccc}
    \toprule
    \multirow{2}{*}{Method} & \multicolumn{3}{c}{AUROC ($\uparrow$)}  & \multicolumn{2}{c}{RMSE ($\downarrow$)} \\ \cmidrule(lr){2-4} \cmidrule(lr){5-6}
    & Diabetes & Ionosphere & Liver & Cholesterol & Yacht \\
    \midrule
    Nonlinear+SVD & 0.6296 & 0.7292 & 0.7032 & \textbf{49.0867} & 15.0002 \\
    \ksvd & \textbf{0.7607} & \textbf{0.8374} & \textbf{0.7100} & 49.1592 & \textbf{14.6489} \\
    \bottomrule
    \end{tabular}
\end{table}

This experiment shows the additional benefit brought by our construction with two data sources and with the asymmetric kernel trick, which does not simply apply SVD to nonlinear features of the input data.
The results are in line with Table \ref{tab:uci} in the main paper and also with the external evaluation of the KL divergence between the densities of the two data sources, where the performance advantage is more evident in Diabetes, Ionosphere, and Yacht, while Liver and Cholesterol show less improvement from considering the proposed asymmetric construction.
In fact, our experiments show that \ksvd is an effective tool to learn more informative features when the given data present asymmetric properties and it also shows overall better performance or comparable results for general datasets; we do not aim to claim that the asymmetric construction is always essential, but rather it is task-dependent and provides more possibilities for practitioners.

\subsection{More results on the asymmetric \nystrom method}

\begin{figure}[H]
    \centering
    \begin{subfigure}[b]{0.3\textwidth}
        \centering
        \includegraphics{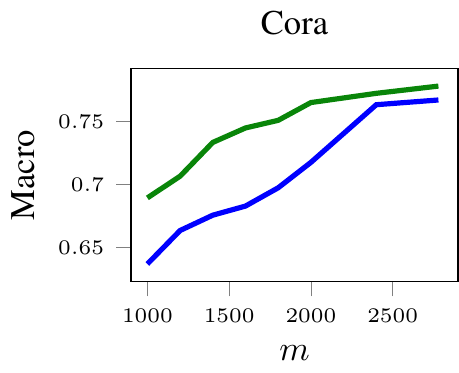}
        \label{fig:nystrom:macro:cora}
    \end{subfigure}
    \hfill
    \begin{subfigure}[b]{0.3\textwidth}
        \centering
        \includegraphics{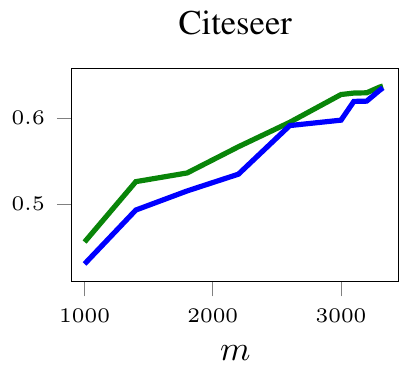}
        \label{fig:nystrom:macro:citeseer}
    \end{subfigure}
    \hfill
    \begin{subfigure}[b]{0.3\textwidth}
        \centering
        \includegraphics{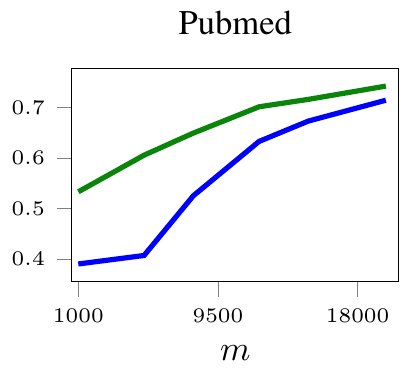}
        \label{fig:nystrom:macro:pubmed}
    \end{subfigure}

    \bigskip
    \begin{subfigure}[b]{0.3\textwidth}
        \centering
        \includegraphics{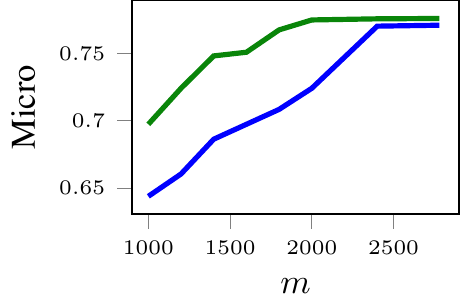}
        \label{fig:nystrom:micro:cora}
    \end{subfigure}
    \hfill
    \begin{subfigure}[b]{0.3\textwidth}
        \centering
        \includegraphics{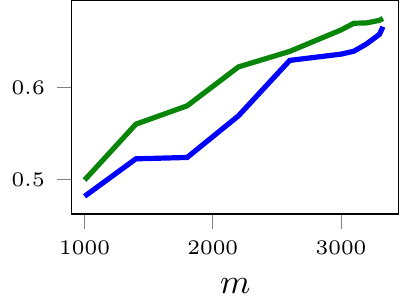}
        \label{fig:nystrom:micro:citeseer}
    \end{subfigure}
    \hfill
    \begin{subfigure}[b]{0.3\textwidth}
        \centering
        \includegraphics{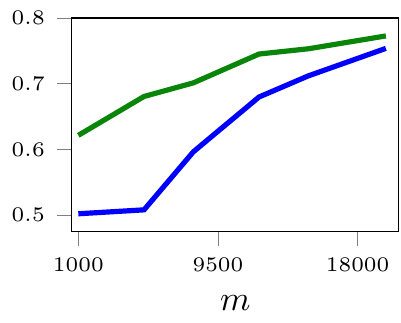}
        \label{fig:nystrom:micro:pubmed}
    \end{subfigure}
    \caption{F1 Score (both Micro and Macro) at different subsamplings $m$ with the asymmetric and symmetric \nystrom method. Green line: \ksvd, blue line: KPCA.}
    \label{fig:nystrom}
\end{figure}

In Fig.~\ref{fig:nystrom}, the node classification F1 score is reported for multiple number of subsamplings $m$, where \ksvd (green line) employs the asymmetric Nystr\"om method and KPCA (blue line) uses the symmetric Nystr\"om, both employing the RBF kernel. 
Note that, as explained in Section \ref{sec:general:data} in the main paper, the resulting kernel matrix  $G$ in  \ksvd  maintains the asymmetry even with the (symmetric) RBF function, as the kernel is applied to two different data sources $\mathcal X$ and $\mathcal Z$.
Note that the data matrix is square, so we can set $m=n$ for the subsamplings of the asymmetric \nystrom.
The proposed asymmetric method \ksvd shows superior performance at all considered $m$ compared to KPCA without significant decrease in accuracy of the solution due to the subsampling.

\pgfplotsset{
    NystromF1/.style={
        width=0.99\linewidth,
        xlabel=$m$,
        height=3.75cm,
        legend pos=south east,
        legend cell align={left},
        xtick pos=left, ytick pos=left,
        xtick align=outside, ytick align=outside,
        every axis plot/.append style={no marks, line width=1.5pt},
        every axis legend/.code={\let\addlegendentry\relax},
        every tick label/.append style={font=\tiny},
    },
    NystromF1Cora/.style={
        NystromF1,
        xmin=900, xmax=2900,
    },
    NystromF1Citeseer/.style={
        NystromF1,
        xmin=900, xmax=3450,
    },
    NystromF1Pubmed/.style={
        NystromF1,
        xmin=600, xmax=20500,
        xtick={1000,9500,18000},
        scaled x ticks=false
    },
}

\section{Detailed experimental setups}

\subsection{Feature learning experiments}
In all experiments, we conduct 10-fold cross validation for determining kernel hyperparameters with grid searches in the same range for fair comparisons. The employed nonlinear kernels in the experiments are $\hat{\kappa}_{\rm{RBF}}(\bm{x},\bm{z}) = \exp(-\frac{\|\bm{x}-\bm{z}\|_2^2}{\gamma^2})$ and $\kappa_{\rm{SNE}}(\bm{x},\bm{z}) = \frac{\exp(-\|\bm{x}-\bm{z}\|_2^2/\gamma^2)}{\sum_{\bm{z}\in \mathcal{Z}}\exp(-\|\bm{x}-\bm{z}\|_2^2/\gamma^2)}$ with hyperparameter $\gamma$.
In the node classification experiments, we denote $X=[\bm x_1, \dots, \bm x_N]^\top$ as the asymmetric adjacency matrix with $X_{ij}$ as the directed similarity between node $i$ and node $j$. 
KPCA is conducted for feature extraction in the following way: we compute symmetric kernel matrix $\hat{K}$ s.t. $\hat{K}_{ij}=\hat{k}(\bm x_i, \bm x_j)$, with (symmetric) RBF kernel $\hat{k}$, and its top $r$ eigenvectors are taken as the extracted features taken as input to the LSSVM classifier. 
PCA is conducted similarly by taking the linear kernel $\hat{k}(\bm x_i, \bm x_j)=\bm x_i^\top \bm x_j$. 
For all methods, we employ an LSSVM classifier with regularization parameter set to 1 and we utilize the one-vs-rest scheme. 
In the graph reconstruction task, with the  feature embeddings extracted by all tested methods, we recover the matrix that reflects the edges between nodes and then the connections between  each node. For a given node $v$ with the out-degree $k_v$, the  closest $k_v$ nodes to $v$ in feature space are searched to reconstruct the adjacency matrix. The $\ell_1, \ell_2$ norms between $X$ and its reconstruction are evaluated.

We give the details of the employed graphs below.

\begin{table}[ht]
    \small
    \centering
    \caption{Descriptions of the tested directed graph datasets.} 
    \label{tab:graph:data}
    \begin{tabular}{lccc}
    \toprule
    {Datasets} & {Cora} & {Citeseer} & {Pubmed} \\
    \midrule
    \# Classes & 7 & 6 & 3   \\
    \# Nodes & 2078 & 3327 & 19717 \\
    \# Edges & 5429 & 4732& 44338   \\ 
    \bottomrule
    \end{tabular}
\end{table}

\subsection{\nystrom experiments}
In this part, we evaluate the efficiency of the proposed asymmetric Nystr\"om method with comparisons to other standard solvers. 
We compare our method with three common SVD solvers: truncated SVD (SVD) from the ARPACK library, Symmetric \nystrom \cite{NIPS2000_19de10ad} applied to $GG^\top$ and $G^\top G$, and randomized SVD (RSVD) \citep{halko2011}.
We employ the Lanczos Method at rank $r$ \citep{lehoucq1998} for the SVD subproblem of Symmetric \nystrom, and
we employ RSVD at rank $r$ for the SVD subproblem of asymmetric \nystrom.
Truncated SVD is run to machine precision for comparison.
For a given tolerance $\varepsilon$, we stop training when $\eta < \varepsilon$, with $\eta$ being the accuracy of a solution.
In particular, for RSVD, we increase the number of oversamples until the target tolerance is reached.
For the \nystrom methods, we increase the number of subsamples $m$ until the target tolerance is reached.
We use random subsampling for all \nystrom methods.
The kernel is chosen to be the SNE kernel with bandwidth set as $\gamma=k\sqrt{M\gamma_x}$, with $\gamma_x$ the variance of the training data and data-dependent $k$ ($k=1$ for Cora and Citeseer, $k=0.5$ for Pubmed).

In the experiments of Figure \ref{fig:nystrom} in Supplementary Material and of Fig. \ref{fig:nystrom:cora} in the main body, we compare the node classification performance of KPCA using Symmetric \nystrom against \ksvd using our asymmetric \nystrom. We use the RBF kernel for both KPCA and \ksvd, with $\gamma$ tuned via 10-fold cross validation. Note that \ksvd achieves higher performance at all considered subsamplings $m$, even if both methods use the RBF kernel. 
Similarly, even when symmetric kernel functions are chosen, the resulting $G$ matrix in the \ksvd solution  \eqref{eq:ksvd} still maintains the asymmetry, as the kernel is applied to two different data sources $\mathcal X$ and $\mathcal Z$.

\section{Algorithm}
Algorithm \ref{alg:ksvd} details the \ksvd algorithm for feature extraction given an arbitrary matrix $A$. The features extracted by the \ksvd algorithm consist of both the left and right singular vectors of an asymmetric matrix kernel $G$. Note that, contrary to KPCA, the left and right singular vectors of $G$ do not coincide, as $G$ is asymmetric.

\begin{algorithm}[H] 
	\caption{\ksvd algorithm.}\label{alg:ksvd}
\begin{algorithmic}[1]
		\Function{\ksvd}{$A \in \mathbb{R}^{N \times M}$, $r\leq \min \{N, M\}$}
        \State Get the two data sources $\mathcal{X} =\{\bm x_i\triangleq  A[i,:] \}_{i=1}^N$ and $\mathcal{Z}=\{\bm z_j\triangleq  A[:,j] \}_{j=1}^M$
        \State Get the transformation matrix $C$ by Remark \ref{rmk:c:matrix} in the main paper
        \State Choose kernel function $\kappa$  and  compute the asymmetric kernel  $G$ with $G_{ij} = \kappa(C^\top \bm x_i, \bm z_j)$
        \State Center $G$ matrix
        \State Compute the SVD of $G=USV^\top$ 
        \State Preserve the first $r$ singular values/vectors in $\tilde{U}, \tilde{S}, \tilde{V}$
        \State
        \Return $\tilde{U}, \Lambda, \tilde{V}$
        \EndFunction		
	\end{algorithmic}
\end{algorithm}

Algorithm \ref{alg:compat} details the realization of the compatibility matrix discussed in Remark \ref{rmk:c:matrix} in the main paper. 
Below, we consider the case $M > N$, where we construct the projection matrix $C_x \in \mathbb{R}^{M \times N}$ such that $XC_x \in \mathbb{R}^{N \times N}$. 
If $N > M$, we rather construct $C_z \in \mathbb{R}^{N \times M}$ such that $ZC_z \in \mathbb{R}^{M \times M}$. 
The construction of $C_z$ mirrors the algorithm for $C_x$ with the appropriate changes.
In the case of square matrix with $N=M$, $C=I_N$, with $I_N$ the identity matrix of size $N \times N$.

\begin{algorithm}[H]
    \caption{Compatibility Matrix Realization.} \label{alg:compat}
	\begin{algorithmic}[1]
		\Function{Compatibility}{$\mathcal{X}=\{\bm x_i \in \mathbb{R}^M \}_{i=1}^N$}
        \State Define $X = [\bm x_1, \dots, \bm x_N]^\top$
        \If{projection on $\bm x_i$}
        \State $C_x = \arg \min \limits_{ C} \norm{ X - X  C C^\top}^2_{\rm F}$  \Comment{Alternative $a_1$}
        \ElsIf{randomized projection} 
        \State $C_x = \text{randn}(M, N)$ \Comment{Alternative $a_2$}
        \ElsIf{pseudoinverse}
        \State $C_x =\left((XX^\top)^\dag X \right)^\top$  \Comment{Alternative $a_0$}
        \EndIf
        \State
        \Return $C_x$
        \EndFunction		
	\end{algorithmic}
\end{algorithm}

\section{Proof of Proposition \ref{prop:dual}}
\begin{proof}
The Lagrangian of \eqref{eq:ksvd} writes
\begin{align*}
L(B_\phi,B_\psi,\Lambda) &= \frac{1}{2}||G^\top B_\phi||^2_F + \frac{1}{2}||G B_\psi||^2_F - \Tr((B_\phi^\top G B_\psi - I_d)\Lambda),
\end{align*}
with Lagrange multipliers $\Lambda$. The KKT conditions are
\begin{align}
\frac{\partial L}{\partial B_\psi} &= G^\top G B_\psi - G^\top B_\phi \Lambda=0 \label{eq:kkt:v} \\
\frac{\partial L}{\partial B_\phi} &= G G^\top B_\phi - G B_\psi \Lambda=0 \label{eq:kkt:u} \\
\frac{\partial L}{\partial \Lambda} &= B_\phi^\top G B_\psi - I_d=0. \label{eq:kkt:lambda}
\end{align}
Note that the last KKT condition \eqref{eq:kkt:lambda} recovers the orthogonality constraints and has the form of a diagonalization of $G$. 
Let the (compact) singular value decomposition of matrix $G$ be $G=\sum_{l=1}^r d_l \bm u_l \bm v_l^\top$, with $\bm u_l^\top \bm u_l = \bm v_l^\top \bm v_l = 1$ for $l=1,\dots,r$ and $d_1 \geq \dots \geq d_r > 0$.
Therefore, with the scaling $\bm b_{\phi_l} \triangleq \frac{1}{\sqrt{d_l}} \bm u_l$ and $\bm b_{\psi_l} \triangleq \frac{1}{\sqrt{d_l}} \bm v_l$, we obtain $\bm b_{\phi_i}^\top G \bm b_{\psi_j} = \delta_{ij}$ satisfying \eqref{eq:kkt:lambda}. Note that this scaling does not change the solutions of \eqref{eq:kkt:v} and \eqref{eq:kkt:u}. In matrix form, with $B_\phi=[\bm b_{\phi_1}, \dots,  \bm b_{\phi_r}], B_\psi=[\bm b_{\psi_1},\dots, \bm b_{\psi_r}], U=[\bm u_1, \dots,  \bm u_r], V=[\bm v_1,\dots, \bm v_r]$, and diagonal matrix $D=\text{diag}(d_1,\dots,d_r)$, we can write the solution as
\begin{align}
B_\phi &= U D^{-1/2} \\
B_\psi &= V D^{-1/2} \\
\Lambda &= D.
\end{align}
Equations \eqref{eq:kkt:v} and \eqref{eq:kkt:u} are satisfied, as
\begin{align}
VDU^\top UDV^\top VD^{-1/2} - VDU^\top UD^{-1/2}D = VD^{-3/2}-VD^{-3/2}=0 \\
UDV^\top VDU^\top UD^{-1/2} - UDV^\top VD^{-1/2}D = UD^{-3/2}-UD^{-3/2}=0.
\end{align}
\end{proof}

\section{Broader impacts and limitations}

\paragraph{Broader impact}
The goal of this paper is to investigate KSVD from the perspective of two covariances across  feature spaces and the joint pursuit of maximal variances. Unlike that classical KPCA resorting to symmetry, KSVD extends SVD to nonlinearity and meanwhile pertains the exploration of asymmetry, which could  benefit researchers and practitioners working with feature learning on  general data, especially the directed data. We also connect with the work from Schmidt on a pair of adjoint eigenfunctions with an asymmetric kernel and analogously derive the asymmetric \nystrom method that can facilitate the efficiency.  Hopefully, our work could bring some new insights and attract attention in the community for more generic feature learning tools   with considerations to the asymmetry residing in the data. Thus, by far, we have  seen  nearly  no negative societal impacts.

\paragraph{Open problems and possible future works} 
With the asymmetric feature learning in KSVD, 
more in-depth knowledge discovery  can be investigated, such as the interpretation on how much information that can be gained from the asymmetry and the application to the bi-clustering task that simultaneously clusters the row data and column data.  Though provided with the asymmetric \nystrom, more sophisticated sampling techniques can be specified to maintain less performance drop for tackling very large-scale data. Correspondingly, compared to the fruitful theoretical analysis on  Mercer kernels, e.g., KPCA, more analytical results can follow up on the extended asymmetric kernel learning.
These are possible directions for future work.

\end{document}